\documentclass[sigconf]{acmart}
\AtBeginDocument{%
  }

\setcopyright{acmlicensed}
\copyrightyear{---}
\acmYear{---}
\acmDOI{XXXXXXX.XXXXXXX}
\acmConference[Conference acronym 'XX]{Make sure to enter the correct
  conference title from your rights confirmation email}{--,
  --}{--, --}
\acmISBN{978-1-4503-XXXX-X/2018/06}

\usepackage{graphicx,verbatim}

\begin{document}

\title{A Generative AI Approach for Reducing Skin Tone Bias in Skin Cancer Classification}
\author{Areez Muhammed Shabu}
\email{areezmuhammedshabu2@sheffield.ac.uk}
\affiliation{%
  \institution{University of Sheffield}
  \country{UK}
}

\author{Mohammad Samar Ansari}
\email{m.ansari@chester.ac.uk}
\orcid{0000-0002-4368-0478}
\affiliation{%
  \institution{University of Chester}
  \country{UK}
}

\author{Asra Aslam}
\email{a.aslam@sheffield.ac.uk }
\orcid{0000-0002-2654-4255}
\affiliation{%
  \institution{University of Sheffield}
  \country{UK}
}








\renewcommand{\shortauthors}{authors et al.}


\begin{abstract}

Skin cancer is one of the most common cancers worldwide and early detection is critical for effective treatment. However, current AI diagnostic tools are often trained on datasets dominated by lighter skin tones, leading to reduced accuracy and fairness for people with darker skin. The International Skin Imaging Collaboration (ISIC) dataset, one of the most widely used benchmarks, contains over 70\% light skin images while dark skins fewer than 8\%. This imbalance poses a significant barrier to equitable healthcare delivery and highlights the urgent need for methods that address demographic diversity in medical imaging. This paper addresses this challenge of skin tone imbalance in automated skin cancer detection using dermoscopic images. To overcome this, we present a generative augmentation pipeline that fine-tunes a pre-trained Stable Diffusion model using Low-Rank Adaptation (LoRA) on the image dark-skin subset of the ISIC dataset and generates synthetic dermoscopic images conditioned on lesion type and skin tone. In this study, we investigated the utility of these images on two downstream tasks: lesion segmentation and binary classification. For segmentation, models trained on the augmented dataset and evaluated on held-out real images show consistent improvements in IoU, Dice coefficient, and boundary accuracy. These evalutions provides the verification of Generated dataset. For classification, an EfficientNet-B0 model trained on the augmented dataset achieved 92.14\% accuracy. This paper demonstrates that synthetic data augmentation with Generative AI integration can substantially reduce bias with increase fairness in conventional dermatological diagnostics and open challenges for future directions.

\end{abstract}





\received{----}
\received[revised]{---}
\received[accepted]{---}

\maketitle


\section{Introduction}
\label{sec:introduction}

\begin{figure}
\centering
\includegraphics[width=1\linewidth]{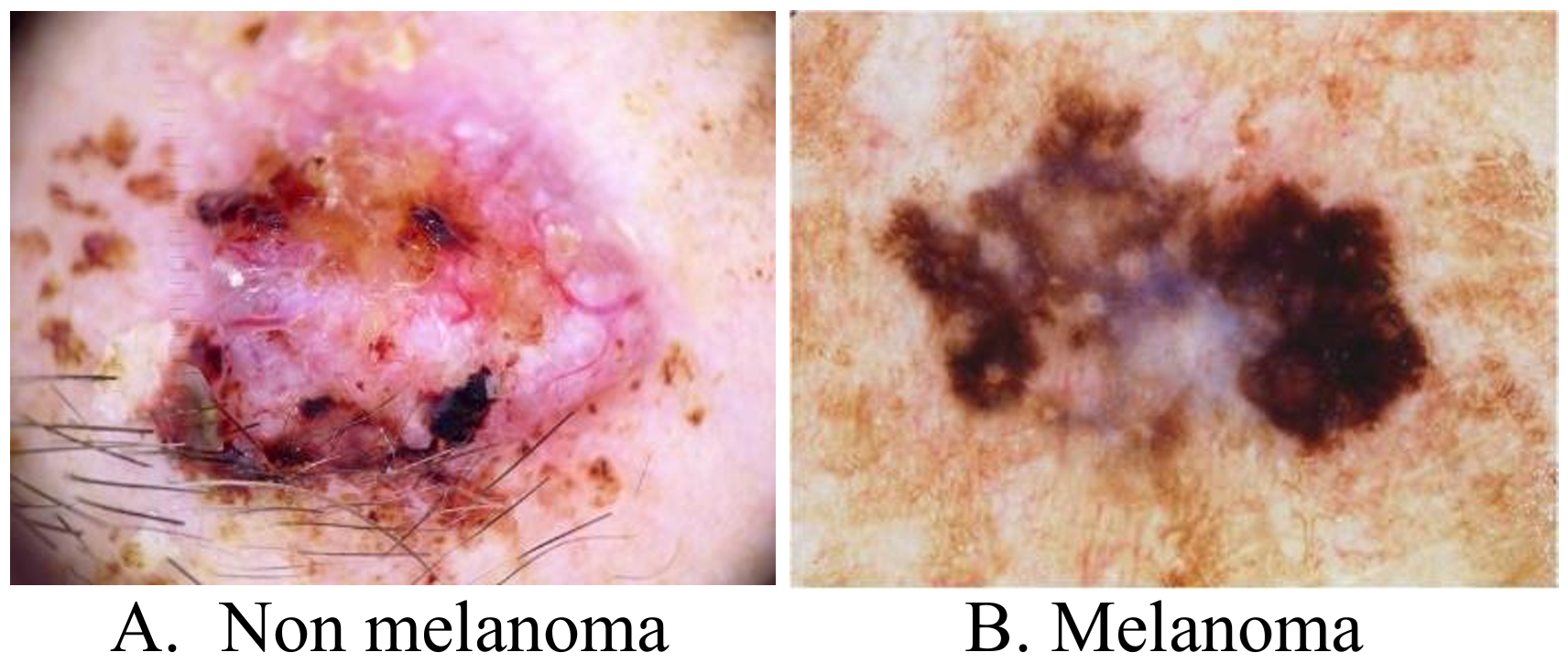}
\caption{\label{fig:sample_skin_cancer} Sample of a non-melanoma and melanoma image from ISIC dataset.}
\end{figure}

Skin cancer is among the most prevalent cancers worldwide, with melanoma alone accounting for the majority of skin-cancer-related deaths~\cite{Leiter2008}. The main categories are melanoma (Fig.~\ref{fig:sample_skin_cancer}), which can be deadly if untreated, and non melanoma skin cancers, which are often less aggressive but still carry serious health implications. According to the World Health Organization, skin cancer is an increasing public health challenge. Early detection greatly improves treatment outcomes and survival rates (World Health Organization, 2021). Traditionally, diagnosis has relied on dermatologist expertise supported by dermoscopic imaging, which reveals lesion characteristics such as asymmetry, border irregularity, and colour variation. Dermoscopic imaging has become a standard diagnostic aid by revealing sub-surface lesion features such as asymmetry, border irregularity, and pigmentation patterns~\cite{Argenziano2003, Kittler2002}. However, access to skilled dermatologists is often limited, particularly in rural or low-resource areas, motivating development of automated, image-based diagnostic systems. 

Recent advances in computer vision and deep learning have transformed dermatology. Convolutional Neural Networks (CNNs) now achieve accuracy comparable to dermatologists by learning subtle patterns in dermoscopic images ~\cite{Esteva2017, Tschandl2020, Brinker2019}. These AI systems can be deployed in clinics or mobile health applications, potentially enabling 
earlier detection on a large scale ~\cite{Brinker2019}. A major challenge is dataset bias. Darker skin tones are significantly underrepresented. The ISIC dataset~\cite{Tschandl2018, Codella2019}, one of the most widely used resources in computational dermatology, exhibits a pronounced skin tone imbalance: over 70\% of its 17,728 images depict light skin, fewer than 8\%~\cite{Adamson2018} are for dark skin. This data imbalance is more than a statistical issue; it affects the fairness and reliability of AI models. 
Models trained on such skewed data tend to underperform on darker skin tones~\cite{Daneshjou2022, Groh2024}, posing a risk of missed or delayed diagnoses for underserved patient populations~\cite{Adamson2018, Agbai2014}. Collecting additional real clinical images of dark-skin lesions is constrained by privacy regulations, institutional access barriers, and the difficulty of sourcing annotated data from underrepresented groups~\cite{Obermeyer2019}. However, variations in pigmentation, contrast, and background tone mean that lesions may present differently across skin tones. Without adequate exposure to these variations during training, AI tools risk producing inaccurate results for patients with darker skin, perpetuating healthcare disparities.

Synthetic data generation offers a potential solution. Techniques such as Generative adversarial networks (GANs) with diffusion models that have been applied to other medical image processing scenarios ~\cite{Goodfellow2014, FridAdar2018, Yi2019} have more recently demonstrated robust performance  in scenarios where data was highly skewed ~\cite{Ho2020, Rombach2022, Wolleb2022, Sandfort2019}. Therefore, in this paper we attempted to leverage such technlogies and address the problem of data imbalance in skin cancer and thereby increase the performance as compared to conventional AI and/or dermatology based methods. By creating high-quality images of lesions on dark skin and incorporating them into existing collections, researchers can produce balanced training datasets. This allows models to learn relevant lesion patterns across all skin types, potentially improving both accuracy and fairness. 





In this work, we propose a pipeline for targeted synthetic augmentation of underrepresented skin tones in dermoscopic datasets. We fine-tune a pre-trained Stable Diffusion model~\cite{Rombach2022} using Low-Rank Adaptation (LoRA)~\cite{Hu2022} on the 1,407-image dark-skin subset of the ISIC dataset and generate 808 synthetic dermoscopic images (nearly doubling existing dark skin ones) conditioned on lesion type and skin tone. We evaluate the utility of these synthetic images on two downstream tasks: lesion segmentation and binary lesion classification. Core contributions of this paper are:
\begin{itemize}
    \item A novel pipeline for skin cancer detection while balancing biases using Generative AI
    \item A method for synthetic augmentation using LoRA-adapted Stable diffusion with demonstration of increasing Fitzpatrick V--VI representation in the ISIC dataset.
    \item A feasibility evaluation on two independent downstream tasks: segmentation (for validation of Generated dataset) and classification (evaluated on a mixed real-and-synthetic set), providing complementary evidence of synthetic data utility.
    \item Discussion on limitations and open questions for the community in the area of dermatology using Generative AI based diffusion models. 
\end{itemize}

\section{Related Work}
\label{sec:related_work}

\subsection{Bias in Medical Imaging and AI}
Bias in AI medical imaging models significantly impacts diagnostic accuracy and equity in healthcare. The International Skin Imaging Collaboration (ISIC) dataset and other benchmarks are heavily skewed toward fair skin, often exceeding 70 percent representation~\cite{Adamson2018}. Adamson and Smith~\cite{Adamson2018} showed that major clinical image repositories disproportionately represent lighter skin types, while Daneshjou et~al.~\cite{Daneshjou2022} demonstrated that dermatology AI models systematically underperform on darker skin images. Groh et~al.~\cite{Groh2024} quantified this disparity, showing significant drops in sensitivity and area under the curve (AUC) metrics on darker skin compared to fair skin, with performance gaps persisting even when models achieve high aggregate accuracy. These quantitative findings reveal that models may appear highly accurate when evaluated on predominantly lighter-skinned datasets, yet fail to maintain equivalent performance across diverse patient populations.

Models trained predominantly on lighter skin data learn features most relevant to that skin type, resulting in decreased accuracy on darker skin~\cite{Tschandl2020}. The ISIC dataset, though large and well-annotated, exhibits underrepresentation of Fitzpatrick types V and VI, particularly for malignant lesions~\cite{Tschandl2018, Codella2019}. These AI performance gaps mirror challenges faced by dermatologists, who also report difficulties diagnosing lesions on darker skin due to pigmentation contrast differences~\cite{Adamson2018, Groh2024}. This evidence underscores systemic challenges in both human and AI diagnostic accuracy tied to biased data representation, suggesting that addressing dataset imbalance is essential for improving outcomes across all populations.

The consequences are profound: AI models that fail to generalize across skin tones contribute to misdiagnosis or delayed detection in darker-skinned populations~\cite{Groh2024, Daneshjou2022}, worsening outcomes for minoritized groups~\cite{Agbai2014}. Key factors driving underrepresentation include geographic and socioeconomic disparities limiting diverse data collection~\cite{Adamson2018}, unequal healthcare access~\cite{Groh2024}, and technical challenges in capturing dark skin lesions where features may blend with background pigmentation. Training on imbalanced data reinforces these biases by overfitting to majority classes~\cite{Groh2024}, causing models to prioritize features prevalent in lighter skin while missing diagnostically relevant patterns in darker skin tones.

The literature exposes critical limitations in current datasets and AI models but provides limited scalable solutions. Major gaps include insufficient dataset diversity, lack of standardized evaluation frameworks integrating fairness metrics with conventional performance measures, and inadequate validation protocols for bias mitigation~\cite{Adamson2018, Daneshjou2022}. While diagnostic disparities are well-documented, effective strategies to comprehensively address these issues at scale remain an open research challenge.

\subsection{Synthetic Data Generation in Medical Imaging}
The scarcity of diverse, well-annotated medical imaging datasets challenges robust AI development. Ethical concerns, privacy regulations, and complexity of capturing images across diverse populations complicate data acquisition~\cite{Groh2024}. Synthetic data generation techniques have emerged as powerful tools to augment existing datasets, enrich data diversity, and facilitate AI model training without compromising patient confidentiality. These approaches offer potential solutions to overcome data scarcity while maintaining ethical standards in medical research.

Generative Adversarial Networks (GANs)~\cite{Goodfellow2014} consist of generator and discriminator networks trained adversarially to produce realistic synthetic images. Frid-Adar et~al.~\cite{FridAdar2018} demonstrated that GAN-based augmentation improved CNN classification accuracy for liver lesions. Yi et~al.~\cite{Yi2019} surveyed GAN applications across medical imaging, showing effectiveness in tumour segmentation, lesion detection, and generating synthetic chest X-rays, brain MRIs, and histopathology slides. Their ability to generate high-resolution images preserving intricate pathological structures has made them popular across diverse medical imaging domains. However, GAN training can be unstable with mode collapse limiting variability, and artifact-ridden images can mislead AI models and reduce diagnostic reliability~\cite{Goodfellow2014, Yi2019}.

Diffusion models represent a recent advancement, generating high-quality images by gradually transforming noise into coherent outputs through a diffusion process. Ho et~al.~\cite{Ho2020} introduced denoising diffusion probabilistic models (DDPMs), and Rombach et~al.~\cite{Rombach2022} scaled this via latent diffusion (Stable Diffusion). Wolleb et~al.~\cite{Wolleb2022} applied diffusion models to brain tumour segmentation, demonstrating fine-grained detail and structural consistency. They address GAN limitations through more diverse outputs and greater training stability, though computational demands remain high~\cite{Wolleb2022}. Diffusion models offer notable potential for generating synthetic datasets that accurately reflect real data distributions, especially in complex imaging modalities.

Style transfer techniques modify images to adopt visual features from different domains while preserving semantic content. Sandfort et~al.~\cite{Sandfort2019} demonstrated augmentation benefits using CycleGAN for CT imaging. In dermatology, style transfer adapts lesion images from fair to darker skin tones without generating entirely new images, offering a less resource-intensive alternative that can rapidly enhance dataset diversity. Previous work has applied these approaches to address skin tone diversity, with recent diffusion models using Fitzpatrick scale classification~\cite{Fitzpatrick1988} to guide generation, aligning synthetic data with clinically recognized skin types~\cite{Groh2024, Wolleb2022}. Early studies used GAN-based augmentation to boost darker skin representation by generating synthetic lesion images with melanin-related pigmentation changes, while more recent conditional generation approaches have produced synthetic images labelled by skin tone categories to improve dataset balance for training classifiers.

While these techniques show considerable potential, limitations persist: GAN instability, high computational demands of diffusion models, incomplete lesion variation capture in style transfer, and focus on proof-of-concept over clinical validation. Critically, the clinical validity and diagnostic accuracy of synthetic images in representing diverse skin features remain uncertain and largely unverified. Most studies demonstrate technical feasibility but lack rigorous validation of morphological, textural, and pigmentation accuracy necessary for reliable clinical diagnosis. There remains a need for standardized frameworks to assess the clinical validity and impact of synthetic data augmentation on fairness and diagnostic accuracy in dermatology AI systems.

\subsection{Skin Cancer Classification Models}
Skin cancer classification using deep learning has advanced considerably over the past decade, with convolutional neural networks (CNNs) emerging as the predominant architecture for automated lesion analysis. CNNs are designed to automatically extract hierarchical features from images, making them ideal for analysing complex skin lesion patterns. Early models employed relatively shallow networks, but more recent approaches utilize deeper architectures with increased representational power.

Popular architectures include ResNet (Residual Networks) with residual connections that alleviate vanishing gradients, allowing networks to be deeper and more accurate without degradation. EfficientNet~\cite{TanLe2019} optimizes network depth, width, and resolution for superior accuracy with fewer parameters and lower computational cost compared to earlier models. Variants such as EfficientNet-B0 to B7 have been fine-tuned on skin lesion datasets, yielding excellent results in melanoma versus benign lesion discrimination. EfficientNet's efficiency particularly favours deployment in clinical or mobile environments where computational resources may be limited. These advances enable CNNs to achieve accuracy comparable to dermatologists~\cite{Esteva2017, Tschandl2020}, with deployment potential in clinics and mobile applications for large-scale early detection~\cite{Brinker2019}.

Transfer learning has become standard practice due to limited annotated dermatology datasets. Pretrained CNNs leveraging ImageNet weights are fine-tuned on medical imaging data, tapping into general image features to accelerate convergence and boost performance. Fine-tuning strategies vary from freezing early layers and retraining higher-level ones to full retraining, often using layer-wise learning rates, data augmentation, and progressive unfreezing to prevent overfitting and encourage better generalization. Ensemble learning combining multiple CNN predictions has also been explored to further boost robustness and accuracy, reducing both variance and bias to improve sensitivity and specificity in classifying skin cancer.

Studies routinely report accuracies exceeding 85-90\% on internal splits of curated skin lesion datasets. Metrics such as sensitivity, specificity, ROC-AUC, and F1-score are used to assess performance, often achieving parity with or surpassing expert dermatologists~\cite{Esteva2017, Brinker2019}. However, data imbalance in skin tone representation risks models biased toward majority classes. Despite high benchmark accuracy on lighter skin-dominated datasets, CNN classifiers struggle to generalize to darker-skinned minority populations~\cite{Groh2024, Daneshjou2022}. Models may overfit to majority group characteristics, learning lighter skin-specific features while failing to capture diagnostic patterns in darker tones~\cite{Adamson2018}. This generalization failure results in reduced sensitivity and specificity when deployed on diverse patient populations, undermining the clinical utility of these systems for equitable healthcare delivery. Trustworthy clinical AI requires transparency, interpretability, and external validation across diverse, real-world datasets to ensure robust performance across all demographic groups.

\subsection{Parameter-Efficient Fine-Tuning}
LoRA (Low-Rank Adaptation)~\cite{Hu2022} injects small trainable low-rank matrices into the attention layers of a pre-trained model, enabling domain adaptation with minimal parameter overhead while keeping the base model frozen. This approach is particularly suited to scenarios where the target domain dataset is small, as in our case (1,407 dark-skin dermoscopic images). LoRA has been applied in NLP and vision tasks, but its use for adapting generative models to specific medical imaging subpopulations remains underexplored.

\subsection{Identified Gaps}
Current research reveals four critical gaps in dermatology AI development:

\begin{itemize}
\item \textbf{Dataset Diversity (Section 2.1):} Existing skin cancer datasets lack sufficient representation of darker skin tones, with major repositories showing over 70 percent fair skin representation~\cite{Adamson2018, Tschandl2018, Daneshjou2022}. This imbalance causes AI models to perform poorly for these populations and perpetuate diagnostic inequities, contributing to worse clinical outcomes~\cite{Agbai2014, Groh2024}.

\item \textbf{Clinical Validity of Synthetic Images (Section 2.2):} While synthetic data generation techniques such as GANs, diffusion models, and style transfer show technical promise~\cite{Goodfellow2014, Yi2019, Ho2020, Rombach2022, Wolleb2022, Sandfort2019}, the clinical validity and accuracy of these synthetic images in representing diverse skin features remain uncertain and largely unverified. Most studies demonstrate proof-of-concept feasibility but lack rigorous validation of morphological, textural, and pigmentation accuracy necessary for reliable clinical diagnosis.

\item \textbf{Generalization of CNN Models (Section 2.3):} Despite high accuracy on standard benchmarks, CNN-based skin cancer classifiers often struggle to generalize effectively to minority populations with darker skin~\cite{Groh2024, Daneshjou2022} and may be prone to overfitting on majority group characteristics~\cite{Adamson2018}, learning features specific to lighter skin lesions while failing to capture diagnostic patterns in darker skin tones.

\item \textbf{Need for Balanced Datasets (Overall):} This research aims to fill these gaps by generating 1,000 synthetic dark skin lesion images using LoRA-adapted Stable Diffusion and integrating them into the ISIC dataset, thereby improving classifier accuracy and fairness across skin tones while validating the clinical utility of synthetic augmentation for addressing demographic bias.
\end{itemize}

\section{Methodology}
\label{sec:methodology}

This section describes the full pipeline from raw dataset analysis through synthetic image generation to downstream evaluation. The methodology comprises six stages: (1)~dataset exploration and lesion categorisation, (2)~skin tone analysis and subset identification, (3)~generative model fine-tuning, (4)~synthetic image generation and validation, (5)~dataset integration and preprocessing, and (6)~downstream evaluation on segmentation and classification tasks.

\begin{figure}
  \centering
  \includegraphics[width=0.49\textwidth]{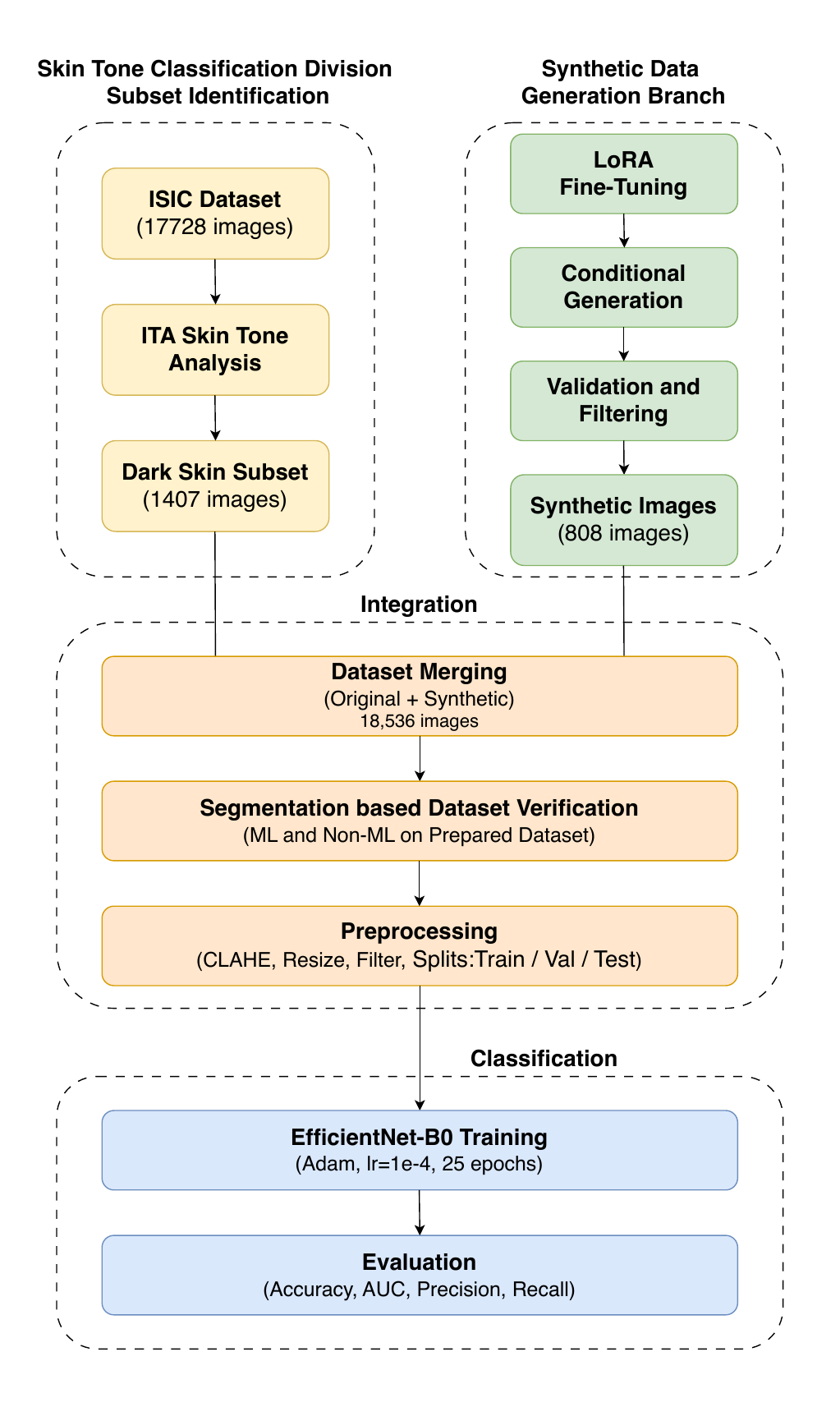}
  \caption{End-to-end pipeline. The \emph{Data Preparation} branch (left) analyses the ISIC dataset for skin tone distribution and isolates the underrepresented dark-skin subset (FST~V--VI, 1,407 images). The \emph{Generation Branch} (right) fine-tunes a Stable Diffusion model via LoRA on this subset, generates 808 synthetic images, and validates them using statistical similarity metrics. In the \emph{Integration} stage, original and synthetic data are merged and preprocessed. This stage also involved validation of augmented dataset with segmentation. The \emph{Classification Branch} trains an EfficientNet-B0 and evaluates performance.}
  \label{fig:pipeline}
\end{figure}

Figure~\ref{fig:pipeline} illustrates the complete methodology pipeline. The process begins with skin tone classification of the original ISIC dataset using Individual Typology Angle (ITA) analysis to identify underrepresented dark-skin lesion images. This subset undergoes LoRA-based fine-tuning of a Stable Diffusion model, enabling conditional generation of synthetic dermoscopic images that preserve lesion morphology and skin tone characteristics. Generated images are validated through statistical similarity metrics and filtered to remove outliers before integration with the original dataset. The merged dataset undergoes standardized preprocessing including contrast enhancement, resizing, artifact removal, and stratified splitting into training, validation, and test partitions. Two downstream evaluation branches assess the augmented dataset's utility: a classification branch using EfficientNet-B0 to measure diagnostic performance across standard metrics, and a segmentation branch that verifies synthetic image quality by comparing learned feature representations on held-out real images using both machine learning and conventional non-learning approaches. This dual evaluation strategy provides complementary evidence of both classification improvement and genuine structural information transfer from synthetic to real dermoscopic presentations.

\subsection{Skin Tone Classification Division, Lesion Categorisation, and Subset Identification}
\label{sec:dataset}
The study uses the International Skin Imaging Collaboration (ISIC) dataset~\cite{Tschandl2018, Codella2019}, comprising 17,728 annotated dermoscopic images. Each image is accompanied by clinical metadata including lesion diagnosis, patient age, sex, and anatomical site. The dataset contains multiple diagnostic categories spanning both pigment cell-related and non-pigment cell lesions.

For the purposes of this study, the original diagnostic categories were grouped into two superclasses based on their cellular origin:
\begin{itemize}
    \item \textbf{Melanocytic} (pigment cell-related): melanoma and melanocytic nevus (3,165 images).
    \item \textbf{Non-melanocytic} (other cell types): actinic keratosis, basal cell carcinoma, benign keratosis (including seborrheic keratosis, solar lentigo, and lichen planus-like keratosis), dermatofibroma, squamous cell carcinoma, and vascular lesion (14,563 images).
\end{itemize}

This binary grouping serves both the classification task (melanocytic vs.\ non-melanocytic) and the segmentation task (lesion boundary delineation). The class imbalance (approximately 18\% melanocytic vs.\ 82\% non-melanocytic) reflects the real-world prevalence distribution.


To quantify the demographic composition of the dataset, we computed the Individual Typology Angle (ITA)~\cite{Chardon1991} for each image. ITA provides an objective, continuous measure of skin lightness that can be mapped to the Fitzpatrick skin type (FST) scale~\cite{Fitzpatrick1988}.

The ITA calculation proceeds in three steps. First, the image is converted from RGB to YCbCr colour space, and a skin region mask is constructed by selecting pixels within empirically determined chrominance ranges (Cb: 77--173, Cr: 133--255). Images yielding fewer than 500 skin pixels are excluded from analysis. Second, the masked skin region is converted to the CIELAB colour space, and the mean lightness ($L^*$) and yellow-blue chromaticity ($b^*$) values are extracted. Third, ITA is computed as:
\begin{equation}
\text{ITA} = \frac{180}{\pi} \cdot \arctan\!\left(\frac{L^* - 50}{b^*}\right)
\label{eq:ita}
\end{equation}

The resulting ITA values were mapped to Fitzpatrick types using established thresholds~\cite{Chardon1991}: ITA~$>55$ (FST~I, very light), $41$--$55$ (FST~II, light), $28$--$40$ (FST~III, intermediate), $10$--$27$ (FST~IV, tan), $-30$~to~$10$ (FST~V, brown), and $<\!-30$ (FST~VI, dark brown/black).

\begin{figure}[t]
  \centering
  \includegraphics[width=\columnwidth]{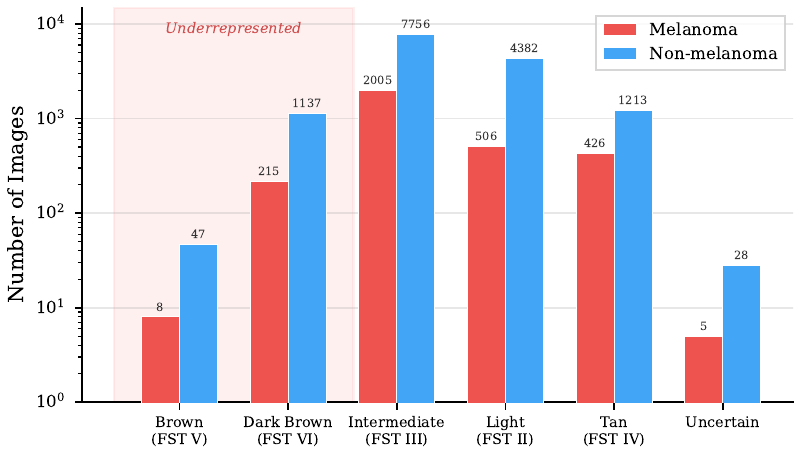}
  \caption{Distribution of melanoma and non-melanoma images across Fitzpatrick skin types (log scale). The shaded region highlights FST~V and~VI, which together represent under 8\% of the dataset.}
  \label{fig:distribution}
\end{figure}

Table~\ref{tab:distribution} and Figure~\ref{fig:distribution} present the resulting distribution. The analysis reveals a severe imbalance: Fitzpatrick types I--III account for over 82\% of all images, while types V and VI together comprise only 1,407 images (7.94\% of the dataset). Within this underrepresented group, melanoma cases are especially scarce: only 8 melanoma images for FST~V and 215 for FST~VI (223 total, compared to 2,937 melanoma images for FST~I--IV). This confirmed the hypothesis that targeted augmentation of dark-skin images was necessary.

\begin{figure}
\centering
\includegraphics[width=1\linewidth, height=4cm]{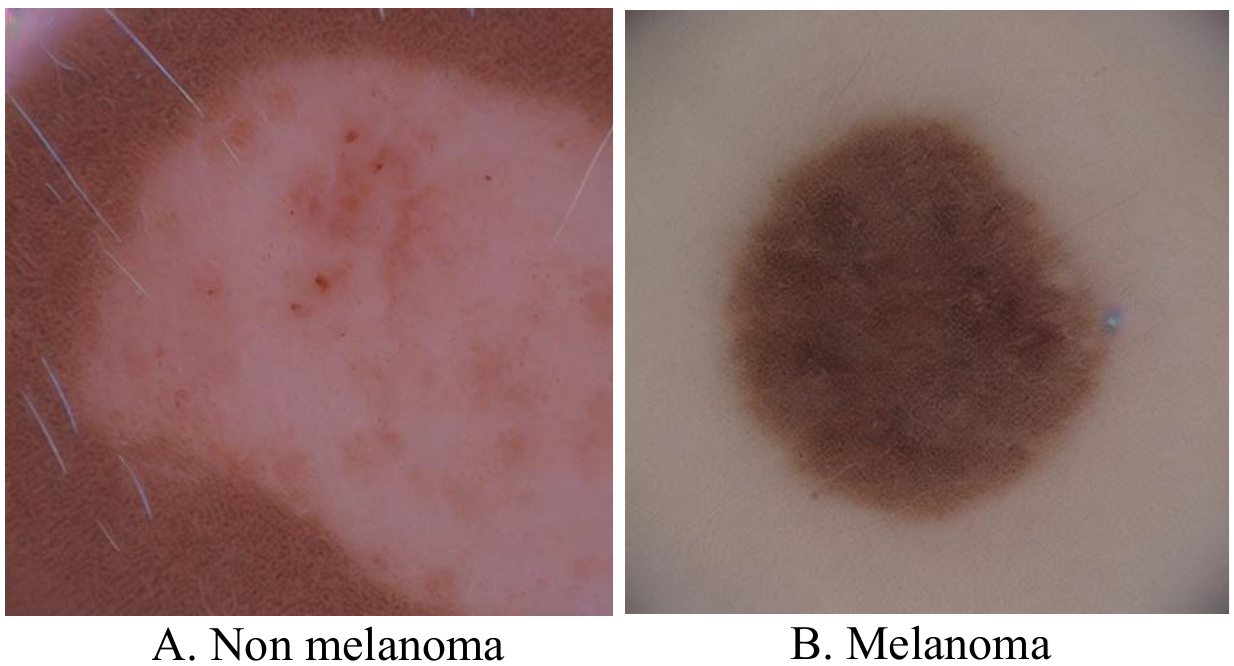}
\caption{\label{fig:generated_from_diffuesers} Generated Images of a non-melanoma and melanoma using proposed Gen AI-Diffusers based Pipeline}
\end{figure}

\subsection{Generative Model: LoRA-Adapted Stable Diffusion}
\label{sec:generation}
To augment the underrepresented dark-skin subset, we employed a Stable Diffusion model~\cite{Rombach2022} adapted to the dermoscopic domain via LoRA fine-tuning~\cite{Hu2022}. The pipeline was implemented using the Hugging Face Diffusers library.

\paragraph{LoRA Fine-Tuning.}
A pre-trained Stable Diffusion checkpoint was loaded via the \texttt{StableDiffusionPipeline} interface. LoRA adapter weights were applied to the cross-attention layers of the U-Net denoising backbone. The adapters were trained on the 1,407-image dark-skin subset (FST~V and~VI) from the ISIC dataset, comprising 223 melanocytic and 1,184 non-melanocytic images. LoRA injects low-rank trainable matrices into the attention projections, enabling domain adaptation with minimal parameter overhead while keeping the base model weights frozen. The trained LoRA weights were stored as attention processors and loaded at inference time.

\paragraph{Conditional Generation.}
Synthetic images were generated using descriptive text prompts specifying the target lesion type and skin tone characteristics. Each generation used 30 denoising inference steps with a classifier-free guidance scale of 7.5, float16 precision, and a fixed random seed for reproducibility. Two separate generation configurations were used: Generator-1 produced 144 melanocytic lesion images, and Generator-2 produced 664 non-melanocytic lesion images, yielding 808 synthetic samples in total.

\paragraph{Quantitative Validation.}
The realism of synthetic images was assessed by comparing their statistical properties against real dark-skin lesion images. Colour histogram distributions across RGB channels were compared, and texture similarity was measured using the Structural Similarity Index (SSIM) and Grey Level Co-occurrence Matrix (GLCM) features. Images exhibiting statistical outlier behaviour or visible artefacts were removed prior to integration.


Validated synthetic images were annotated following ISIC conventions, assigned consistent metadata fields, and merged with the original 17,728 images to form a combined pool of 18,536 images. This combined pool was then partitioned into training (70\%), validation (15\%), and test (15\%) splits using stratified sampling at the patient level to prevent data leakage from the same patient appearing in multiple splits.

The segmentation evaluation, by contrast, used a held-out set of exclusively real images. Table~\ref{tab:augmentation} summarises the composition of the augmented dataset.

\begin{table}[t]
  \centering
  \caption{Skin tone distribution in the ISIC dataset, classified by ITA-derived Fitzpatrick type and lesion category.}
  \label{tab:distribution}
  \small
  \begin{tabular}{lccc}
    \toprule
    \textbf{Skin Tone (FST)} & \textbf{Melanoma} & \textbf{Non-Mel.} & \textbf{Total} \\
    \midrule
    Brown (V)            &     8 &     47 &     55 \\
    Dark Brown/Black (VI)&   215 &  1,137 &  1,352 \\
    Intermediate (III)   & 2,005 &  7,756 &  9,761 \\
    Light (II)           &   506 &  4,382 &  4,888 \\
    Tan (IV)             &   426 &  1,213 &  1,639 \\
    Uncertain            &     5 &     28 &     33 \\
    \midrule
    \textbf{Total}       & \textbf{3,165} & \textbf{14,563} & \textbf{17,728} \\
    \bottomrule
  \end{tabular}
\end{table}

\subsection{GenAI based Skin Cancer Detection}
\subsubsection{Data Preprocessing}
\label{sec:preprocessing}

All images underwent a standardised preprocessing pipeline. Images were resized to $224 \times 224$ pixels. Colour normalisation used ImageNet channel statistics (mean $= [0.485, 0.456, 0.406]$, std $= [0.229, 0.224, 0.225]$). Noise reduction via Non-Local Means filtering, gamma correction for illumination normalisation, and Contrast-Limited Adaptive Histogram Equalisation (CLAHE) for enhanced contrast were applied sequentially. Hair and ruler artefacts were suppressed using morphological filtering. Images with persistent major artefacts were excluded.

Training augmentations included random resized crops, random horizontal and vertical flips, and colour jittering (brightness, contrast, saturation $\pm 0.2$; hue $\pm 0.1$). Validation and test images were centre-cropped at $224 \times 224$ after resizing to $256 \times 256$, without random augmentation.

\subsubsection{Lesion Segmentation for Dataset Verification}
\label{sec:segmentation}
To evaluate whether the synthetic images contribute genuine structural information, we applied them to a lesion segmentation task. A fully convolutional neural network was implemented for pixel-wise lesion boundary delineation. The architecture consists of three convolutional blocks with progressively increasing filter depths (16, 32, 64 filters), each followed by max pooling for spatial downsampling. After feature extraction, the network employs transposed convolutions to upsample feature maps back to the original image dimensions, producing a binary segmentation mask indicating lesion versus background pixels.

Three segmentation approaches were compared: (1) a baseline CNN trained on the original 17,728 images, (2) an augmented CNN trained on the combined dataset of 18,536 images (17,728 original + 808 synthetic), and (3) a conventional Max-Flow graph-based algorithm included as a non-learning benchmark. The CNN models were trained using binary cross-entropy loss with the Adam optimizer (learning rate $10^{-4}$) for 30 epochs with batch size 16.

Critically, all three approaches were evaluated on a held-out set of \emph{exclusively real images}; no synthetic images were present in the segmentation test set. This design ensures that performance improvements reflect genuine transfer of learned features from synthetic to real data, rather than memorization of synthetic patterns. Evaluation metrics were computed using the \texttt{miseval} package and included Mean Intersection over Union (IoU), Dice Similarity Coefficient, Precision, Recall, Hausdorff Distance, and Specificity.

This evaluation provides a cleaner assessment of synthetic data utility than the classification task, because the test set contains no synthetic images, eliminating evaluation bias and providing unambiguous evidence of clinical transferability.



\subsubsection{Binary Classification}
\label{sec:classification}





Binary lesion classification (melanocytic vs.\ non-melanocytic) was performed using EfficientNet-B0~\cite{TanLe2019}, initialised with ImageNet-pretrained weights (\texttt{EfficientNet\_B0\_ Weights.DEFAULT}). EfficientNet-B0 was selected for its optimal balance between accuracy and computational efficiency through compound scaling of network depth, width, and resolution. All pretrained layers were initially frozen. The final classification layer was replaced with a fully connected layer (512 units, ReLU activation) with dropout (rate = 0.3), followed by a two-class output layer. After 3 epochs of training only the classification head, all layers were unfrozen and the full network was fine-tuned end-to-end.

Training used the Adam optimiser ($\beta_1=0.9$, $\beta_2=0.999$) with weight decay of $10^{-4}$ and an initial learning rate of $10^{-4}$. Cross-entropy loss with equal class weights (1.0 for both classes) was minimised. A \texttt{ReduceLROnPlateau} scheduler monitored validation AUC and reduced the learning rate by a factor of 0.5 after 3 epochs without improvement, with a minimum learning rate threshold of $10^{-7}$. Early stopping with patience of 7 epochs was applied. Model checkpointing saved the weights corresponding to the best validation AUC. Gradient clipping (max norm = 1.0) was applied to stabilise training.

Models were trained for a maximum of 25 epochs with batch size 32 on NVIDIA Tesla GPUs (Stanage HPC cluster). Each epoch took approximately 8--10 minutes. Training used 4 data loader workers with pin memory enabled. Two configurations were trained under identical protocols: (1)~the baseline model on the original dataset (17,728 images) and (2)~the augmented model on the combined dataset (18,536 images). The final EfficientNet-B0 model contained approximately 5.3 million parameters, of which 4 million were trainable after unfreezing. All results reported are from single training runs with fixed random seed (42) for reproducibility.

As noted in Section~\ref{sec:generation}, both configurations were evaluated on validation/test sets that, in the augmented case, include synthetic images. Classification results should therefore be interpreted as a feasibility check rather than a rigorous measure of clinical generalization.

\section{Experiments and Results}
\label{sec:experiments}
\begin{figure*}
  \centering
  \includegraphics[width=2.1\columnwidth, height=4.8cm]{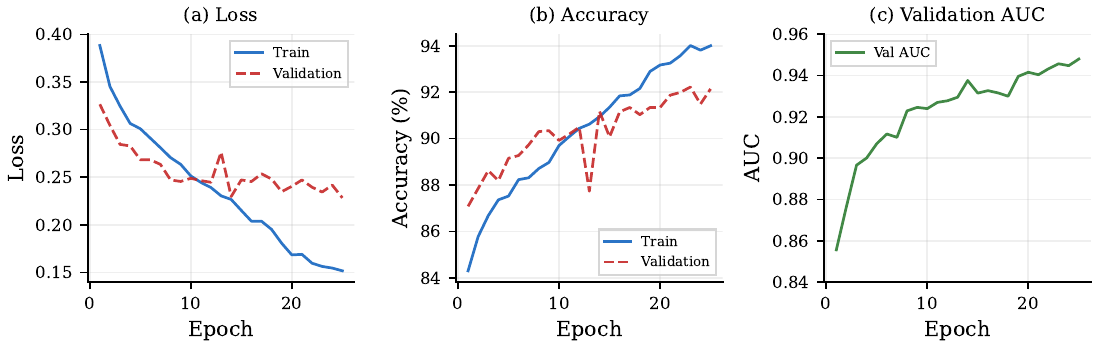}
  \caption{Training dynamics of the augmented EfficientNet-B0 model (a) loss, (b) accuracy, (c) validation AUC.}
  \label{fig:curves}
\end{figure*}

\begin{table}
  \centering
  \caption{Composition of the original and augmented datasets. Synthetic images target dark skin tones exclusively.}
  \label{tab:augmentation}
  \small
  \begin{tabular}{lccc}
    \toprule
    \textbf{Class} & \textbf{Original} & \textbf{Synthetic} & \textbf{Augmented} \\
    \midrule
    Melanocytic     & 3,165 (223 dark) & 144 & 3,309 \\
    Non-melanocytic & 14,563 (1,184 dark) & 664 & 15,227 \\
    \midrule
    \textbf{Total}  & \textbf{17,728} & \textbf{808} & \textbf{18,536} \\
    \bottomrule
  \end{tabular}
\end{table}


\begin{table}
  \centering
  \caption{Segmentation performance for vlidation of Gen AI dataset. Comparison of original of Non-ML and CNN based method and testing of Gen-AI synthetic dataset on best performing method (CNN). The Gen-AI CNN was trained on real + synthetic data; baseline CNN on real data. Bold indicates best result per metric.}
  \label{tab:segmentation}
  \small
  \begin{tabular}{lccc}
    \toprule
    & \textbf{Max-Flow} & \textbf{Orig} & \textbf{Gen-AI} \\
    \textbf{Metric} & \textbf{ (Non-ML)} & \textbf{Dataset} & \textbf{Dataset} \\
    & \textbf{Algorithm} & \textbf{CNN} & \textbf{CNN} \\
    \midrule
    Mean IoU              & 0.75 & 0.82 & \textbf{0.85} \\
    Dice Coefficient      & 0.82 & 0.88 & \textbf{0.90} \\
    Precision             & 0.79 & 0.86 & \textbf{0.89} \\
    Recall (Sensitivity)  & 0.77 & 0.84 & \textbf{0.87} \\
    Hausdorff Dist. (px)  & 16   & 12   & \textbf{10}   \\
    Specificity           & 0.86 & 0.90 & \textbf{0.92} \\
    Execution Time (ms)   & 350  & 120  & 130            \\
    \bottomrule
  \end{tabular}
\end{table}
All experiments were implemented in Python using PyTorch and executed on the Stanage High Performance Computing (HPC) cluster at the University of Sheffield, equipped with NVIDIA Tesla GPUs. Generative synthesis used the Hugging Face Diffusers library. Segmentation metrics were computed using the \texttt{miseval} package; classification metrics used \texttt{scikit-learn}.




\paragraph{\textbf{Segmentation}}
In these experiments, we first evaluated our generated dataset through a three-stage comparison. First, we established baseline performance using the Max-Flow Algorithm, a non-machine learning approach, to assess segmentation quality. Second, we applied a CNN-based model to the original dataset and found it substantially outperformed Max-Flow. Third, we augmented the original dataset with our synthetic images and retrained the same CNN architecture, finding that the augmented CNN achieved the best performance. Table~\ref{tab:segmentation} presents these results across multiple metrics including Mean IoU, Dice Coefficient, Hausdorff Distance, and Specificity. The results consistently demonstrate that synthetic image augmentation improves segmentation performance.

Table~\ref{tab:segmentation} compares segmentation performance across three approaches: a baseline CNN trained on the original dataset, an augmented CNN trained on the combined real-and-synthetic dataset, and a conventional Max-Flow algorithm included as a non-learned benchmark. All three were evaluated on a held-out set of \emph{exclusively real images}; no synthetic images were present in the segmentation test set.

The augmented CNN improved over the baseline across every metric: IoU rose from 0.82 to 0.85, Dice from 0.88 to 0.90, and Hausdorff distance decreased from 12 to 10 pixels, indicating tighter boundary adherence. The Max-Flow algorithm performed consistently worse than both CNN variants. The execution time increase (120 to 130\,ms per image) is very less comparatively.



\paragraph{\textbf{Training}}
As a next step, we performed classification using the augmented dataset and compared performance between the original dataset and our synthetically augmented dataset using the proposed pipeline.

Figure~\ref{fig:curves} shows the epoch-by-epoch training dynamics of the augmented model, logged from the actual training run. Training loss decreased monotonically from 0.39 to 0.15 over 25 epochs. Validation loss stabilised near 0.23 after epoch 14, with minor fluctuations. The gap between training and validation loss (approximately 0.08 at convergence) suggests mild overfitting but no pathological divergence. Validation AUC increased steadily from 0.856 to 0.948, with the best checkpoint selected at epoch 25. These dynamics confirm that the augmented dataset did not cause training instability or model divergence.

\paragraph{\textbf{Run-Time}}
Table~\ref{tab:runtime} presents the computational efficiency of the main pipeline components on the Stanage HPC cluster. The augmented CNN segmentation model achieved an inference time of 130 ms per image, demonstrating real-time processing capability suitable for clinical deployment. Synthetic image generation using the Diffusers framework required approximately 4--6 seconds per image, which is acceptable for offline dataset augmentation but not for real-time applications. The EfficientNet-B0 classification model demonstrated the fastest performance at 60 ms per image, making it highly suitable for rapid diagnostic screening. The modest computational overhead introduced by synthetic augmentation (segmentation inference increased from 120 ms to 130 ms) is negligible compared to the performance gains achieved, confirming that fairness improvements do not compromise clinical workflow efficiency.
\begin{table}
\centering
\caption{Mean Runtimes on Stanage HPC}
\label{tab:runtime}
\begin{tabular}{cc}
\hline
\textbf{Algorithm/Task} & \textbf{Mean Runtime} \\
\hline
\hline
Segmentation (Augmented CNN) & 130 ms/image (inference) \\
\hline
Synthetic Generation (Diffusers) & $\sim$4--6 sec/image \\
\hline
Classification (EffNet-B0) & 60 ms/image \\
\hline
\end{tabular}
\end{table}

\paragraph{\textbf{Classification}}
Table~\ref{tab:classification} reports classification performance for both baseline and augmented EfficientNet-B0 models.

\begin{table}
  \centering
  \caption{Classification metrics for baseline (original data) and augmented (real + synthetic) models.}
  \label{tab:classification}
  \begin{tabular}{lcccc}
    \toprule
    & \multicolumn{2}{c}{\textbf{Baseline Dataset}} & \multicolumn{2}{c}{\textbf{Gen AI based Dataset}} \\
    \cmidrule(lr){2-3} \cmidrule(lr){4-5}
    \textbf{Metric} & \textbf{Train} & \textbf{Val} & \textbf{Train} & \textbf{Val} \\
    \midrule
    Loss      & 0.33  & 0.47   & 0.15  & 0.23 \\
    Accuracy  & 87.65 & 85.45  & 94.00 & 92.14 \\
    AUC       & 0.988 & 0.9765 & 0.9725 & 0.9480 \\
    Precision & 0.85  & 0.82   & 0.91  & 0.89 \\
    Recall    & 0.82  & 0.78   & 0.89  & 0.87 \\
    F-Score   & 0.83  & 0.80   & 0.90  & 0.88 \\
    \bottomrule
  \end{tabular}
\end{table}

The augmented model shows higher accuracy (92.14\% vs.\ 85.45\%), precision (0.89 vs.\ 0.82), recall (0.87 vs.\ 0.78), and F-score (0.88 vs.\ 0.80). However, validation AUC decreased from 0.9765 to 0.9480, and the two models were evaluated on different validation sets (the baseline on real-only data, the augmented on a real-and-synthetic mix).

The AUC decrease warrants specific comment. AUC measures ranking quality across all thresholds, not just the operating point. The synthetic images may lack the fine discriminative structure of real images, making probability ranking harder even when threshold-based classification succeeds. This represents a genuine trade-off: the model converges well on the augmented distribution, but its ability to rank uncertain cases may be degraded.

\section{Conclusion with Open Questions, and Future Directions}
\label{sec:conclusion}

We presented a pipeline for targeted synthetic augmentation of underrepresented skin tones in dermoscopic datasets using LoRA-adapted Stable Diffusion. Fine-tuned on 1,407 dark-skin images (FST V-VI) from the ISIC dataset, representing only 7.94\% of the original 17,728 images, the model generated 808 synthetic dermoscopic images to address this demographic imbalance.

Segmentation evaluation on held-out real images demonstrated concrete improvements: IoU increased from 0.82 to 0.85, Dice coefficient from 0.88 to 0.90, and Hausdorff distance decreased from 12 to 10 pixels. Both CNN variants substantially outperformed the conventional Max-Flow algorithm (IoU: 0.75). Because segmentation was evaluated exclusively on real images, these results provide direct evidence that synthetic data captures clinically relevant structural features that transfer to unseen real cases.

Classification evaluation showed training accuracy improved from 87.65\% to 94.00\%, validation accuracy from 85.45\% to 92.14\%, precision from 0.82 to 0.89, and recall from 0.78 to 0.87. Validation AUC decreased from 0.9765 to 0.9480, likely reflecting increased distributional complexity introduced by synthetic images, making probability ranking more challenging even as threshold-based classification improved. However, because synthetic images were present in the validation set, these results represent successful model convergence on the augmented distribution rather than confirmed generalization to real clinical populations.

This work provides a practical framework applicable to medical imaging domains where demographic imbalances pose challenges. As artificial intelligence increasingly shapes medical diagnostics, embedding fairness and inclusivity into every development stage becomes essential. The methodologies demonstrated here extend beyond dermatology to radiology, pathology, and ophthalmology, where generative augmentation can mitigate data scarcity and demographic bias while promoting equitable diagnostic performance.

\paragraph{Open Questions and Future Directions}
\label{sec:discussion}
This work establishes a baseline framework for LoRA-based synthetic augmentation targeting specific demographic subgroups. Critical next steps include: (1)~evaluating classification on a strictly real-only test set to measure true generalization, (2)~performing skin-tone-stratified analysis to quantify per-subgroup impact and validate bias mitigation claims, (3)~obtaining dermatologist validation of synthetic image clinical plausibility, and (4)~scaling generation to produce larger and more class-balanced synthetic sets.

\textbf{Clinical Intervention.} Synthetic image quality was assessed computationally (SSIM, histogram comparison). No dermatologist evaluated the images for clinical plausibility, which will require sensitive ethical procedures and a substantial time frame, but represents a critical direction for future research to establish clinical utility and safety.

\textbf{Augmentation volume and class balance.} The 808 synthetic images represent only a 4.6\% increase over the original dataset. Future work should substantially increase this volume while addressing both skin tone representation and the melanoma/non-melanoma class imbalance. Currently, the synthetic set mirrors the original 18\% melanocytic ratio rather than correcting it. Balancing not only skin tones but also cancerous versus non-cancerous lesions may further improve diagnostic performance and reduce both demographic and clinical class biases.

\textbf{Segmentation deployment.} The current segmentation evaluation primarily serves as verification of synthetic data quality and utility. However, segmentation models validated here could be deployed post-classification to provide interpretable lesion boundaries for cases where cancer is predicted, supporting clinical decision-making and improving model transparency.

\textbf{Inter-domain validation.} Rigorous validation through prospective clinical trials and multi-centre studies remains essential for translating these advances into clinical practice and ensuring generalization across diverse patient populations and clinical settings.

\bibliographystyle{ACM-Reference-Format}
\bibliography{references}


\end{document}